\documentclass[10pt,twocolumn,letterpaper]{article}

\usepackage{cvpr}
\usepackage{times}
\usepackage{epsfig}
\usepackage{graphicx}
\usepackage{amsmath}
\usepackage{amssymb}

\usepackage[breaklinks=true,bookmarks=false]{hyperref}

\cvprfinalcopy 

\def\cvprPaperID{3524} 

\ifcvprfinal\fi

\begin{document}

\title{Arbitrary Shape Scene Text Detection with Adaptive Text Region Representation}
\author{Xiaobing Wang\textsuperscript{1}, Yingying Jiang\textsuperscript{1}, Zhenbo Luo\textsuperscript{1}, Cheng-Lin Liu\textsuperscript{2, 3}, Hyunsoo Choi\textsuperscript{4}, Sungjin Kim\textsuperscript{4}\\
\textsuperscript{1}Samsung Research China - Beijing, Beijing 100028, China\\
\textsuperscript{2}National Laboratory of Pattern Recognition \\ Institute of Automation of Chinese Academy of Sciences, Beijing 100190, China\\
\textsuperscript{3}University of Chinese Academy of Sciences, Beijing 100049, China\\
\textsuperscript{4}Samsung Research, Seoul 06765, Korea\\
{\tt\small \{x0106.wang, yy.jiang, zb.luo, hsu.choi, sj9373.kim\}@samsung.com,} {\tt\small liucl@nlpr.ia.ac.cn}\\
}

\maketitle
\thispagestyle{empty}

\begin{abstract}
   Scene text detection attracts much attention in computer vision, because it can be widely used in many applications such as real-time text translation, automatic information entry, blind person assistance, robot sensing and so on. Though many methods have been proposed for horizontal and oriented texts, detecting irregular shape texts such as curved texts is still a challenging problem. To solve the problem, we propose a robust scene text detection method with adaptive text region representation. Given an input image, a text region proposal network is first used for extracting text proposals. Then, these proposals are verified and refined with a refinement network. Here, recurrent neural network based adaptive text region representation is proposed for text region refinement, where a pair of boundary points are predicted each time step until no new points are found. In this way, text regions of arbitrary shapes are detected and represented with adaptive number of boundary points. This gives more accurate description of text regions. Experimental results on five benchmarks, namely, CTW1500, TotalText, ICDAR2013, ICDAR2015 and MSRA-TD500, show that the proposed method achieves state-of-the-art in scene text detection.
\end{abstract}

\section{Introduction}

Text is the most fundamental medium for communicating semantic information. It appears everywhere in daily life: on street nameplates, store signs, product packages, restaurant menus and so on. Such texts in natural environment are known as scene texts. Automatically detecting and recognizing scene texts can be very rewarding with numerous applications, such as real-time text translation, blind person assistance, shopping, robots, smart cars and education. An end-to-end text recognition system usually consists of two steps: text detection and text recognition. In text detection, text regions are detected and labeled with their bounding boxes. And in text recognition, text information is retrieved from the detected text regions. Text detection is an important step for end-to-end text recognition, without which texts can not be recognized from scene images. Therefore, scene text detection attracts much attention these years.

While traditional optical character reader (OCR) techniques can only deal with texts on printed documents or business cards, scene text detection tries to detect various texts in complex scenes. Due to complex backgrounds and variations of font, size, color, language, illumination condition  and orientation, scene text detection becomes a very challenging task. And its performance was poor when hand designed features and traditional classifiers were used before deep learning methods become popular. However, the performance has been much improved in recent years, significantly benefitted from the development of deep learning. Meanwhile, the research focus of text detection has shifted from horizontal scene texts~\cite{Authors_ICDAR13} to multi-oriented scene texts~\cite{Authors_ICDAR15} and more challenging curved or arbitrary shape scene texts~\cite{Authors_TextSnake}. Therefore, arbitrary shape scene text detection is focused on in this paper.

\begin{figure*}[htbp]
\begin{center}
\includegraphics[width=0.92\linewidth]{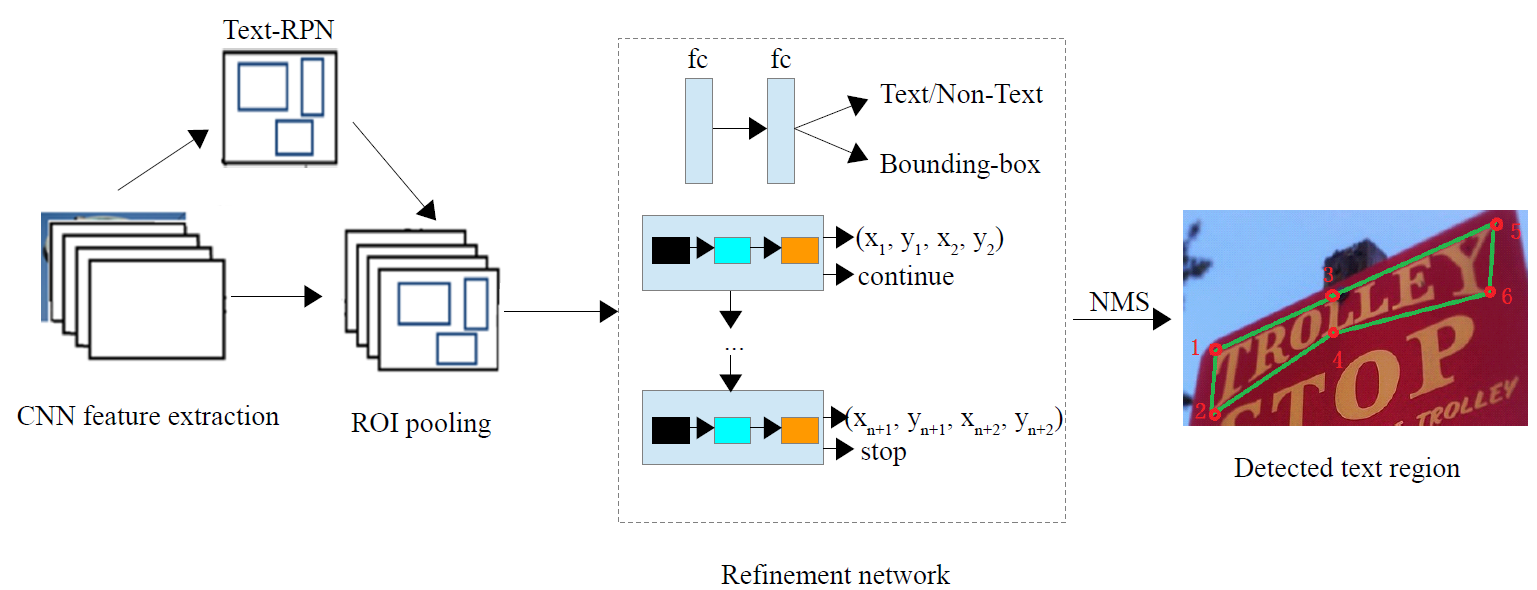}
\end{center}
   \caption{Flowchart of the proposed method for arbitrary shape scene text detection. With adaptive text region representation used, detected text regions can be labeled with adaptive number of pairwise points.}
\label{fig1}
\end{figure*}

In this paper, we propose an arbitrary shape scene text detection method using adaptive text region representation, as shown in Figure~\ref{fig1}. Given an input image, a text region proposal network (Text-RPN) is first used for obtaining text proposals. The Convolutional Neural Network (CNN) feature maps of the input image are also obtained in this step. Then, text proposals are verified and refined with a refinement network, whose input are the text proposal features obtained by using region of interest (ROI) pooling to the CNN feature maps. Here, three branches including text/non-text classification, bounding box refinement and recurrent neural network (RNN) based adaptive text region representation exist in the refinement network. In the RNN, a pair of boundary points are predicted each time step until the stop label is predicted. In this way, arbitrary shape text regions can be represented with adaptive number of boundary points. For performance evaluation, the proposed method is tested on five benchmarks, namely, CTW1500, TotalText, ICDAR2013, ICDAR2015 and MSRA-TD500. Experimental results show that the proposed method can process not only multi-oriented scene texts but also arbitrary shape scene texts including curved texts. Moreover, it achieves state-of-the-art performances on the five datasets.

\section{Related work}
Traditional sliding window based and Connected component (CC) based scene text detection methods had been widely used before deep learning became the most promising machine learning tool. Sliding window based methods~\cite{Authors_textflow,Authors_symline} move a multi-scale window over an image and classify the current patch as text or non-text. CC based methods, especially the Maximally Stable Extremal Regions (MSER) based methods~\cite{Authors_MSER,Authors_MSER2}, get character candidates by extracting CCs. And then, these candidate CCs are classified as text or non-text. These methods usually adopt a bottom-up strategy and often need several steps to detect texts (e.g., character detection, text line construction and text line classification). As each step may lead to misclassification, the performances of these traditional text detection methods are poor.

Recently, deep learning based methods have become popular in scene text detection. These methods can be divided into three groups, including bounding box regression based methods, segmentation based methods, and combined methods. Bounding box regression based methods~\cite{Authors_deepdirect,Authors_R2CNN,Authors_textboxplus,Authors_textbox,Authors_RSR,Authors_DMPnet}, which are inspired by general object detection methods such as SSD~\cite{Authors_SSD} and Faster R-CNN~\cite{Authors_Fasterrcnn}, treat text as a kind of object and directly estimate its bounding box as the detection result. Segmentation based methods~\cite{Authors_pixelink,Authors_TextSnake,Authors_fcn} try to solve the problem by segmenting text regions from the background and an additional step is needed to get the final bounding boxes. Combined methods~\cite{Authors_MaskTextspotter} use a similar strategy as Mask R-CNN~\cite{Authors_Maskrcnn}, in which both segmentation and bounding box regression are used for better performance. However, its processing time is increased because more steps are needed than previous methods. Among the three kinds of methods, bounding box regression based methods are the most popular in scene text detection, benefitted from the development of general object detection.

For bounding box regression based methods, they can be divided into one-stage methods and two-stage methods. One-stage methods including Deep Direct Regression~\cite{Authors_deepdirect}, TextBox~\cite{Authors_textbox}, TextBoxes++~\cite{Authors_textboxplus}, DMPNet~\cite{Authors_DMPnet}, SegLink~\cite{Authors_seglink} and EAST~\cite{Authors_EAST}, directly estimate bounding boxes of text regions in one step. Two-stage methods include R2CNN~\cite{Authors_R2CNN}, RRD~\cite{Authors_RSR}, RRPN~\cite{Authors_RRPN},  IncepText~\cite{Authors_Incepttext} and FEN~\cite{Authors_FENet}. They consist of text proposal generation stage, in which candidate text regions are generated, and bounding box refinement stage, in which candidate text regions are verified and refined to generate the final detection result. Two-stage methods usually achieve higher performances than one-stage methods. Therefore, the idea of two-stage detection is used in this paper.

While most proposed scene text detection methods can only deal with horizontal or oriented texts, detecting arbitrary shape texts such as curved text attracts more attention recently. In CTD~\cite{Authors_ctw1500}, a polygon of fixed 14 points are used to represent text region. Meanwhile, recurrent transverse and longitudinal offset connection (TLOC) is proposed for accurate curved text detection. Though a polygon of fixed 14 points is enough for most text regions, it is not enough for some long curve text lines. Besides, 14 points are too many for most horizontal and oriented texts, while 4 points are enough for these texts. In TextSnake~\cite{Authors_TextSnake}, a text instance is described as a sequence of ordered, overlapping disks centered at symmetric axes of text regions. Each disk is associated with potentially variable radius and orientation, which are estimated via a Fully Convolutional Network (FCN) model. Moreover, Mask TextSpotter~\cite{Authors_MaskTextspotter} which is inspired by Mask R-CNN, can handle text instances of irregular shapes via semantic segmentation. Though TextSnake and Mask TextSpotter both can deal with text of arbitrary shapes, pixel-wise predictions are both needed in them, which need heavy computation.

Considering a polygon of fixed number of points is not suitable for representing text regions of different shapes, an adaptive text region representation using different numbers of points for texts of different shapes is proposed in this paper. Meanwhile, a RNN is employed to learn the adaptive representation of each text region, with which text regions can be directly labeled and pixel-wise segmentation is not needed.

\section{Methodology}
Figure~\ref{fig1} shows the flowchart of the proposed method for arbitrary shape text detection, which is a two-stage detection method. It consists of two steps: text proposal and proposal refinement. In text proposal, a Text-RPN is used to generate text proposals of an input image. Meanwhile, the CNN feature maps of the input image are obtained here, which can be used in the following. Then, text proposals are verified and refined through a refinement network. In this step, text/non-text classification, bounding box regression and RNN based adaptive text region representation are included. Finally, text regions labeled with polygons of adaptive number of points are output as the detection result.

\subsection{Adaptive text region representation}
The existing scene text detection methods use polygons of fixed number of points to represent text regions. For horizontal texts, 2 points (left-top point and bottom-right point) are used to represent the text regions. For multi-oriented texts, the 4 points of their bounding boxes are used to represent these regions. Moreover, for curved texts, 14 points are adopted in CTW1500~\cite{Authors_ctw1500} for text region representation. However, for some very complex scene texts, such as curved long text, even 14 points may be not enough to represent them well. While for most scene texts such as horizontal texts and oriented texts, less than 14 points are enough and using 14 points to represent these text regions is a waste.

\begin{figure}[htbp]
\begin{center}
\includegraphics[width=0.9\linewidth]{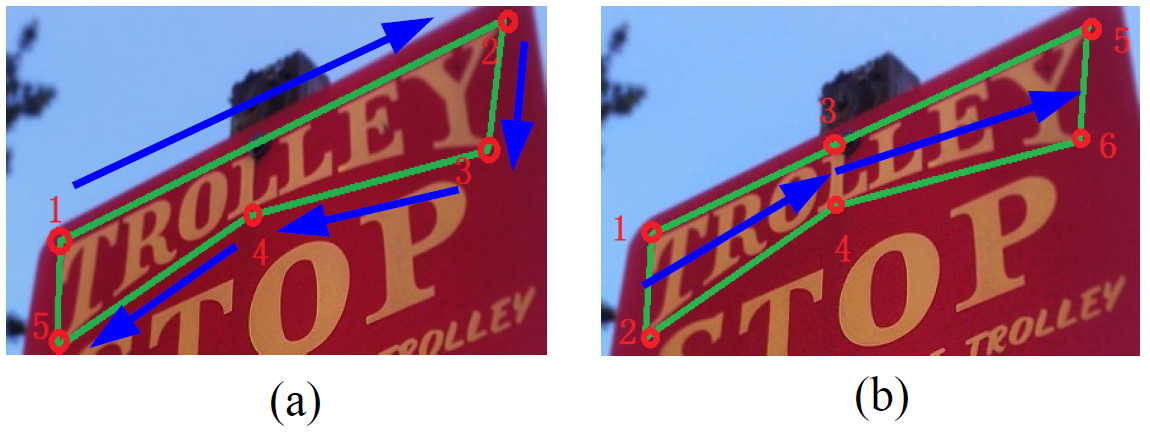}
\end{center}
   \caption{Two methods for adaptive text region representation. (a) text region represented by corner points; (b) text region represented by pairwise points on its top and down boundaries.}
\label{fig7}
\end{figure}

\begin{figure}[htbp]
\begin{center}
\includegraphics[width=\linewidth]{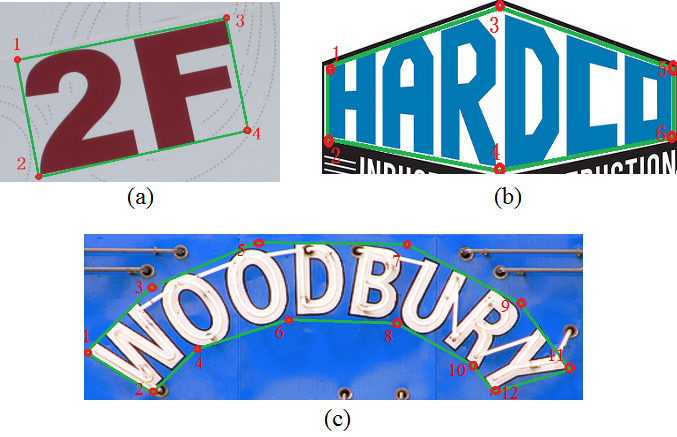}
\end{center}
   \caption{Samples of text regions with adaptive representation. (a) text region represented by 4 points; (b) text region represented by 6 points; (c) text region represented by 12 points.}
\label{fig2}
\end{figure}

Therefore, it is reasonable to consider using polygons of adaptive numbers of points to represent text regions. Easily, we can imagine that corner points on the boundary of a text region can be used for region representation, as shown in Figure~\ref{fig7} (a). And this is similar as the method for annotating general objects~\cite{Authors_polygon}. However, the points in this way are not arranged in a direction and it may be difficult to learn the representation. In the method for annotating general objects, human correction may be needed for accurate segmentation. Considering text regions usually have approximate symmetry top boundary and down boundary as shown in Figure~\ref{fig2}, using the pairwise points from the two boundaries for text region representation may be more suitable. It is much easier to learn the pairwise boundary points from one end to the other end of text region, as shown in Figure~\ref{fig7} (b). In this way, different scene text regions can be represented by different numbers of points precisely, as shown in Figure~\ref{fig2}. Moreover, to our knowledge, we are the first to use adaptive numbers of pairwise points for text region representation.

\subsection{Text proposal}

\begin{figure}[htbp]
\begin{center}
\includegraphics[width=0.95\linewidth]{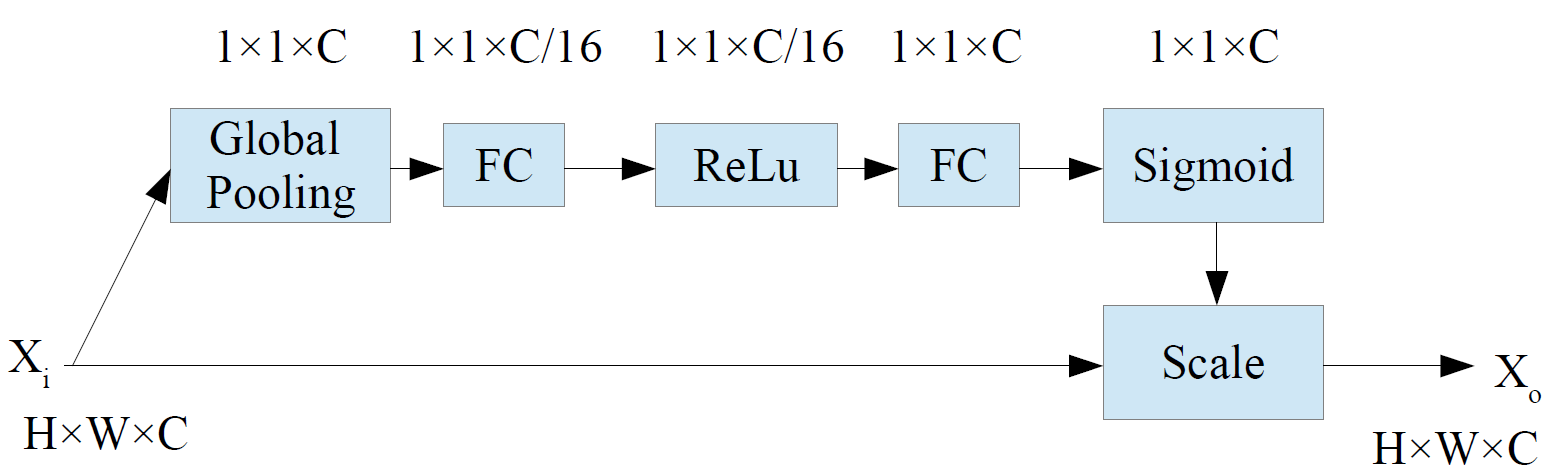}
\end{center}
   \caption{The architecture of SE block.}
\label{fig3}
\end{figure}

When an input image is given, the first step of the proposed method is text proposal, in which text region candidates called text proposals are generated by Text-RPN. The Text-RPN is similar as RPN in Faster R-CNN~\cite{Authors_Fasterrcnn} except different backbone networks and anchor sizes. In the proposed method, the backbone network is SE-VGG16 as shown in Table~\ref{table1}, which is obtained by adding Squeeze-and-Excitation (SE) blocks~\cite{Authors_SEnet} to VGG16~\cite{Authors_VGG}. As shown in Figure~\ref{fig3}, SE blocks adaptively recalibrate channel-wise feature responses by explicitly modelling interdependencies between channels, which can produce significant performance improvement. Here, FC means fully connected layer and ReLU means Rectified Linear Unit function. Moreover, because scene texts usually have different sizes, anchor sizes are set as \{32, 64, 128, 256, 512\} for covering more texts while aspect ratios \{0.5, 1, 2\} are kept.

\begin{table}[htbp]
\begin{center}
\begin{tabular}{|c|c|}
\hline
{\bf Layer} & {\bf Kernel} \\
\hline
Conv1 & $[3 \times 3, 64] \times 2$ \\
Pool1 &  $2 \times 2$, stride 2 \\
SE1 & 4, 64 \\
\hline
Conv2 & $[3 \times 3, 128] \times 2$ \\
Pool2 & $2 \times 2$, stride 2 \\
SE2 & 8, 128 \\
\hline
Conv3 & $[3 \times 3, 256] \times 3$ \\
Pool3 & $2 \times 2$, stride 2 \\
SE3 & 16, 256 \\
\hline
Conv4 & $[3 \times 3, 512] \times 3$ \\
Pool4 & $2 \times 2$, stride 2 \\
SE4 & 32, 512 \\
\hline
Conv5 & $[3 \times 3, 512] \times 3$ \\
Pool5 & $2 \times 2$, stride 2 \\
SE5 & 32, 512 \\
\hline
\end{tabular}
\end{center}
\caption{The architecture of SE-VGG16 network. For SE block, its kernel means the channel numbers of the two FC layers in it.}
\label{table1}
\end{table}

\subsection{Proposal refinement}
After text proposal, text region candidates in the input image are generated, which will be verified and refined in this step. As shown in Figure~\ref{fig1} a refinement network is employed for proposal refinement, which consists of several branches: text/non-text classification, bounding box regression and RNN based adaptive text region representation. Here, text/non-text classification and bounding box regression are similar as other two-stage text detection methods, while the last branch is proposed for arbitrary shape text representation.

For the proposed branch, the input are the features of each text proposal, which are obtained by using ROI pooling to the CNN feature maps generated with SE-VGG16. The output target of this branch is the adaptive number of boundary points for each text region. Because the output length changes for different text regions, it is reasonable to use RNN to predict these points. Therefore, Long Short-Term Memory (LSTM)~\cite{Authors_LSTM} is used here, which is a kind of RNN and popular for processing sequence learning problem, such as machine translation, speech recognition, image caption and text recognition.

\begin{figure}[ht]
\begin{center}
\includegraphics[width=0.75\linewidth]{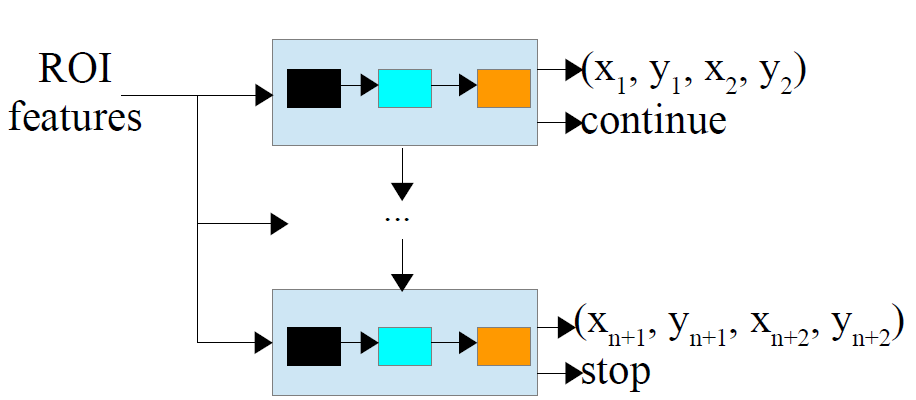}
\end{center}
   \caption{LSTM used for learning text region representation. The input of each time step in the LSTM are the ROI pooling features of the corresponding text proposals.}
\label{fig4}
\end{figure}

Though it is proposed that pairwise boundary points are used for text region representation, different ways can be used for pairwise points representation. Easily, we can imagine that using the coordinates of two pairwise points $(x_i,y_i,x_{i+1},y_{i+1})$ to represent them. In this way, the coordinates of pairwise points are used as the regression targets as shown in Figure~\ref{fig4}. However, pairwise points can be represented in a different way, using the coordinate of their center point $(x_i^c,y_i^c)$, the distance from the center point to them $h_i$, and their orientation $\theta_i$. However, the angle target is not stable in some special situations. For example, angle near $90^{\circ}$ is very similar to angle near $-90^{\circ}$ in spatial, but their angles are quite different. This makes the network hard to learn the angle target well. Besides, the orientation can be represented by $\sin\theta_i$ and $\cos\theta_i$, which can be predicted stably. However, more parameters are needed. Therefore, the coordinates of points $(x_i,y_i,x_{i+1},y_{i+1})$ are used as the regression targets in the proposed method.

The inputs of all time steps in the LSTM used here are the same, which are the ROI pooling features of the corresponding text proposals. And the output of each time step are the coordinates of the pairwise points on text region boundary. Meanwhile, as adaptive numbers of points are used for different text regions, a stop label is needed to represent when the predicting network stops. Because stop label prediction is a classification problem while coordinates prediction is a regression problem, it is not appropriate to put them in the same branch. Therefore, there are two branches in each time step of the LSTM: one for point coordinate regression and one for stop label prediction. At each time step the coordinates of two pairwise boundary points of text region and the label stop/continue are predicted. If the label is continue, the coordinates of another two points and a new label are predicted in the next time step. Otherwise, the prediction stops and text region is represented with the points predicted before. In this way, text regions in the input image can be detected and represented with different polygons made up by the predicted pairwise points.

While Non-Maximum Suppression (NMS) is extensively used to post-process detection candidates by general object detection methods, it is also needed in the proposed method. As the detected text regions are represented with polygons, normal NMS which is computed based on the area of horizontal bounding box is not suitable here. Instead, a polygon NMS is used, which is computed based on the area of the polygon of text region. After NMS, the remaining text regions are output as the detection result.
\subsection{Training objective}
As Text-RPN in the proposed method is similar as the RPN in Faster R-CNN~\cite{Authors_Fasterrcnn}, the training loss of Text-RPN is also computed in the similar way as it. Therefore, in this section, we only focus on the loss function of refinement network in proposal refinement. The loss defined on each proposal is the sum of a text/non-text classification loss, a bounding box regression loss, a boundary points regression loss and a stop/continue label classification loss. The multi-task loss function on each proposal is defined as:

\begin{equation}
		\begin{split}
		L_{sum} =& L_{cls}(p,t) + \lambda_1 t \sum\nolimits_{i\in\{x, y, w, h\}} L_{reg}(v_i,v_i^{\ast})   \\
		&+ \lambda_2 t \sum\nolimits_{i\in\{x_1, y_1, x_2, y_2,\ldots, x_n, y_n\}} L_{reg}(u_i,u_i^{\ast}) \\
        &+ \lambda_3 t \sum\nolimits_{i\in\{l_1, l_2, \ldots, x_{n/2}\}} L_{cls}(l_i,l_i^{\ast}) \\
		\end{split}
	\end{equation}
$\lambda_1$, $\lambda_2$ and $\lambda_3$ are balancing parameters that control the trade-off between these terms and they are set as 1 in the proposed method.

For the text/non-text classification loss term, $t$ is the indicator of the class label. Text is labeled as 1 $(t = 1)$, and background is labeled as 0 $(t = 0)$. The parameter $p = (p_0, p_1)$ is the probability over text and background classes computed after softmax. Then, $L_{cls}(p,t)=-\log p_t$  is the log loss for true class $t$.

For the bounding box regression loss term, $v=(v_x,v_y,v_w,v_h)$ is a tuple of true bounding box regression targets including coordinates of the center point and its width and height, and $v^\ast=(v_x^{\ast},v_y^{\ast},v_w^{\ast},v_h^{\ast} )$ is the predicted tuple for each text proposal. We use the parameterization for $v$ and $v^{\ast}$ given in Faster R-CNN~\cite{Authors_Fasterrcnn}, in which $v$ and $v^{\ast}$ specify scale-invariant translation and log-space height/width shift relative to an object proposal.

For the boundary points regression loss term, $u=(u_{x_1},u_{y_1},\ldots,u_{x_n},u_{y_n})$ is a tuple of true coordinates of boundary points, and $u^\ast=(u_{x_1}^\ast,u_{y_1}^\ast,\ldots,u_{x_n}^\ast,u_{y_n}^\ast )$ is the predicted tuple for the text label. To make the points learned suitable for text of different scales, the learning targets should also be processed to make them scale invariant. The parameters $(u_{x_i}^\ast,u_{y_i}^\ast)$ are processed as following:
\begin{equation}
u_{x_i}^{\ast} = (x_i^\ast - x_a)/w_a, u_{y_i}^\ast = (y_i^\ast - y_a)/h_a,	
\end{equation}
where $x_i^\ast$ and $y_i^\ast$ denote the coordinates of the boundary points, $x_a$ and $y_a$ denote the coordinates of the center point of the corresponding text proposal, $w_a$ and $h_a$ denote the width and height of this proposal.

Let $(w,w^\ast )$ indicates $(v_i,v_i^\ast)$ or $(u_i,u_i^\ast)$, $L_{reg} (w,w^\ast)$ is defined as the smooth L1 loss as in Faster R-CNN~\cite{Authors_Fasterrcnn}:
\begin{equation}
L_{reg}(w,w^\ast) = smooth_{L1}(w - w^\ast),	
\end{equation}
\begin{equation}
smooth_{L1}(x) = \left\{ \begin{array}{rc}	
       0.5x^2 & if |x|<1 \\
       |x|-0.5 & otherwise
       \end{array}\right.
\end{equation}

For the stop/continue label classification loss term, it is also a binary classification and its loss is formatted similar as text/non-text classification loss.

\section{Experiments}

\subsection{Benchmarks}
Five benchmarks are used in this paper for performance evaluation, which are introduced in the following:
\begin{itemize}
\item {\bf CTW1500}: The CTW1500 dataset~\cite{Authors_ctw1500} contains 500 test images and 1000 training images, which contain multi-oriented text, curved text and irregular shape text. Text regions in this dataset are labeled with 14 scene text boundary points at sentence level.
\item {\bf TotalText}: The TotalText dataset~\cite{Authors_totaltext} consists of 300 test images and 1255 training images with more than 3 different text orientations: horizontal, multi-oriented, and curved. The texts in these images are labeled at word level with adaptive number of corner points.
\item {\bf ICDAR2013}: The ICDAR2013 dataset~\cite{Authors_ICDAR13} contains focused scene texts for ICDAR Robust Reading Competition. It includes 233 test images and 229 training images. The scene texts are horizontal and labeled with horizontal bounding boxes made up by 2 points at word level.
\item {\bf ICDAR2015}: The ICDAR2015 dataset~\cite{Authors_ICDAR15} focuses on incidental scene text in ICDAR Robust Reading Competition. It includes 500 testing images and 1000 training images. The scene texts have different orientations, which are labeled with inclined boxes made up by 4 points at word level.
\item {\bf MSRA-TD500}: The MSRA-TD500 dataset~\cite{Authors_swt} contains 200 test images and 300 training images, that contain arbitrarily-oriented texts in both Chinese and English. The texts are labeled with inclined boxes made up by 4 points at sentence level. Some long straight text lines exist in the dataset.

\end{itemize}

The evaluation for text detection follows the ICDAR evaluation protocol in terms of Recall, Precision and Hmean. Recall represents the ratio of the number of correctly detected text regions to the total number of text regions in the dataset while Precision represents the ratio of the number of correctly detected text regions to the total number of detected text regions. Hmean is single measure of quality by combining recall and precision. A detected text region is considered as correct if its overlap with the ground truth text region is larger than a given threshold. The computation of the three evaluation terms is usually different for different datasets. While the results on ICDAR 2013 and ICDAR 2015 can be evaluated through ICDAR robust reading competition platform, the results of the other three datasets can be evaluated with the given evaluation methods corresponding to them.

\subsection{Implementation details}

Our scene text detection network is initialized with pre-trained VGG16 model for ImageNet classification. When the proposed method is tested on the five datasets, different models are used for them, which are trained using only the training images of each dataset with data augmentation. All models are trained $10 \times 10^4$ iterations in total. Learning rates start from $10^{-3}$, and are multiplied by 1/10 after $2 \times 10^4$, $6 \times 10^4$ and $8 \times 10^4$ iterations. We use 0.0005 weight decay and 0.9 momentum. We use multi-scale training, setting the short side of training images as \{400, 600, 720, 1000, 1200\}, while maintaining the long side at 2000.

Because adaptive text region representation is used in the proposed method, it can be simply used for these datasets with text regions labeled with different numbers of points. As ICDAR 2013, ICDAR 2015 and MSRA-TD500 are labeled with quadrilateral boxes, they are easy to be transformed into pairwise points. However, for CTW1500 dataset and TotalText dataset, some operations are needed to transform the ground truthes into the form we needed.

Text regions in CTW1500 are labeled with 14 points, which are needed to be transformed into adaptive number of pairwise points. First, the 14 points are grouped into 7 point pairs. Then, we compute the intersection angle for each point, which is the angle of the two vectors from current point to its nearby two points. And for each point pair, the angle is the smaller one of the two points. Next, point pairs are sorted according to their angles in descending order and we try to remove each point pair in the order. If the ratio of the polygon areas after removing operation to the original area is larger than 0.93, this point pair can be removed. Otherwise, the operation stops and the remaining points are used in the training for text region representation.

Moreover, text regions in TotalText are labeled with adaptive number of points, but these points are not pairwise. For text regions labeled with even number of points, it is easy to process by group them into pairs. For text regions labeled with odd number of points, the start two points and the end two points should be found first, and then the corresponding points to the remaining points are found based on their distances to the start points on the boundary.

The results of the proposed method are obtained on single scale input image with one trained model. Because test image scale has a deep impact on the detection results, such as FOTS~\cite{Authors_FOTS} uses different scales for different datasets, we also use different test scales for different datasets for best performance. In our experiments, the scale for ICDAR 2013 is $960 \times 1400$, the scale for ICDAR 2015 is $1200 \times 2000$ and the scales for other datasets are all $720 \times 1280$.

The proposed method is implemented in Caffe and the experiments are finished using a Nvidia P40 GPU.

\subsection{Ablation study}
In the proposed method the backbone network is SE-VGG16, while VGG16 is usually used by other state-of-the-art methods. To verify the effectiveness of the backbone network, we test the proposed method with different backbone networks (SE-VGG16 vs VGG16) on CTW1500 dataset and ICDAR 2015 dataset as shown in Table~\ref{table2}. The results show that SE-VGG16 is better than VGG16, with which better performances achieved on the two datasets.

\begin{table}[!htbp]
\begin{center}
\begin{tabular}{|c|c|c|c|c|}
\hline
{\bf Backbone} & {\bf Recall} & {\bf Precision} & {\bf Hmean}\\
\hline
\multicolumn{4}{|c|}{CTW1500}\\
\hline
VGG16 & 79.1 & 79.7 & 79.4 \\
\hline
SE-VGG16 & 80.2 & 80.1 & 80.1 \\
\hline
\multicolumn{4}{|c|}{ICDAR2015}\\
\hline
VGG16 & 83.3 & 90.4 & 86.8 \\
\hline
SE-VGG16  & 86.0 & 89.2 & 87.6 \\
\hline
\end{tabular}
\end{center}
\caption{Ablation study on backbone network.}
\label{table2}
\end{table}

Meanwhile, an adaptive text region representation is proposed for text of arbitrary shapes in this paper. To validate its effectiveness for scene text detection, we add an ablation study on text region representation on CTW1500 dataset. For comparison, fixed text region representation directly uses the fixed 14 points as the regression targets in the experiment. Table~\ref{table21} shows the experimental results of different text region representation methods on CTW1500 dataset. The recall of the method with adaptive representation is much higher than fixed representation (80.2\% vs 76.4\%). It justifies that the adaptive text region representation is more suitable for texts of arbitrary shapes.

\begin{table}[!htbp]
\begin{center}
\begin{tabular}{|c|c|c|c|c|}
\hline
{\bf Representation} & {\bf Recall} & {\bf Precision} & {\bf Hmean}\\
\hline
Fixed & 76.4 & 80.0 & 78.2 \\
\hline
Adaptive & 80.2 & 80.1 & 80.1 \\
\hline
\end{tabular}
\end{center}
\caption{Ablation study on text region representation.}
\label{table21}
\end{table}

\subsection{Comparison with State-of-the-arts}
To show the performance of the proposed method for different shape texts, we test it on several benchmarks. We first compare its performance with state-of-the-arts on CTW1500 and TotalText which both contains challenging multi-oriented and curved texts. Then we compare the methods on the two most widely used benchmarks: ICDAR2013 and ICDAR2015. At last we compare them on MSRA-TD500 which contains long straight text lines and multi-language texts (Chinese+English).

Table~\ref{table3} and Table~\ref{table4} compare the proposed method with state-of-the-art methods on CTW1500 and TotalText, respectively. The propose method is much better than all other methods on CTW1500 including the methods designed for curved texts such as CTD, CTD+TLOC and TextSnake (Hmean: 80.1\% vs 69.5\%, 73.4\% and 75.6\%). Meanwhile, it also achieves better performance (Hmean: 78.5\%) than all other methods on TotalText. The performances on the two datasets containing challenging multi-oriented and curved texts mean that the proposed method can detect scene text of arbitrary shapes.

\begin{table}[!htbp]
\begin{center}
\begin{tabular}{|c|c|c|c|}
\hline
{\bf Method} & {\bf Recall} & {\bf Precision} & {\bf Hmean}\\
\hline
SegLink~\cite{Authors_seglink} & 40.0 & 42.3 & 40.8 \\
\hline
EAST~\cite{Authors_EAST} & 49.1 & 78.7 & 60.4 \\
\hline
DMPNet~\cite{Authors_DMPnet} & 56.0 & 69.9 & 62.2 \\
\hline
CTD~\cite{Authors_ctw1500} & 65.2 & 74.3 & 69.5 \\
\hline
CTD+TLOC~\cite{Authors_ctw1500} & 69.8 & 77.4 & 73.4 \\
\hline
TextSnake~\cite{Authors_TextSnake} & {\bf 85.3} & 67.9 &75.6 \\
\hline
{\bf Proposed} & 80.2 & {\bf 80.1} & {\bf 80.1} \\
\hline
\end{tabular}
\end{center}
\caption{Results on CTW1500.}
\label{table3}
\end{table}

\begin{table}[!htbp]
\begin{center}
\begin{tabular}{|c|c|c|c|}
\hline
{\bf Method} & {\bf Recall} & {\bf Precision} & {\bf Hmean}\\
\hline
SegLink~\cite{Authors_seglink} & 23.8 & 30.3 & 26.7 \\
\hline
EAST~\cite{Authors_EAST} & 36.2 & 50.0 & 42.0 \\
\hline
DeconvNet~\cite{Authors_totaltext} & 44.0 & 33.0 & 36.0 \\
\hline
Mask Textspotter~\cite{Authors_MaskTextspotter} & 55.0 & 69.0 & 61.3 \\
\hline
TextSnake~\cite{Authors_TextSnake} & 74.5 & {\bf 82.7} & 78.4 \\
\hline
{\bf Proposed} & {\bf 76.2} & 80.9 & {\bf 78.5} \\
\hline
\end{tabular}
\end{center}
\caption{Results on TotalText.}
\label{table4}
\end{table}

Table~\ref{table5} shows the experimental results on ICDAR2013 dataset. The proposed method achieves the best performance same as Mask Textspotter, whose Hmean both are 91.7\%. Because the proposed method is tested on single scale input image with single model, only the results generated in this situation are used here. The results show that the proposed method can also process horizontal text well.

\begin{table}[!htbp]
\begin{center}
\begin{tabular}{|c|c|c|c|}
\hline
{\bf Method} & {\bf Recall} & {\bf Precision} & {\bf Hmean}\\
\hline
TextBoxes~\cite{Authors_textbox} & 83.0 & 88.0 & 85.0 \\
\hline
SegLink~\cite{Authors_seglink} & 83.0 & 87.7 & 85.3 \\
\hline
He \etal~\cite{Authors_deepdirect} & 81.0 & 92.0 & 86.0 \\
\hline
Lyu \etal~\cite{Authors_coner} & 84.4 & 92.0 & 88.0 \\
\hline
FOTS~\cite{Authors_FOTS} & - & - & 88.2 \\
\hline
RRPN~\cite{Authors_RRPN} & 87.9 & 94.9 & 91.3 \\
\hline
FEN~\cite{Authors_FENet} & 89.1 & 93.6 & 91.3 \\
\hline
Mask Textspotter~\cite{Authors_MaskTextspotter} & 88.6 & {\bf 95.0} & 91.7 \\
\hline
{\bf Proposed} & {\bf 89.7} & 93.7 & {\bf 91.7} \\
\hline
\end{tabular}
\end{center}
\caption{Results on ICDAR2013.}
\label{table5}
\end{table}

Table~\ref{table6} shows the experimental results on ICDAR 2015 dataset and the proposed method achieve the second best performance, which is only a little lower than FOTS  (Hmean: 87.6\% vs 88.0\%). While FOTS is trained end-to-end by combining text detection and recognition, the proposed method is only trained for text detection, which is much easier to train than FOTS. And the results tested on single scale input image with single mode are used here. The results show that the proposed method achieves comparable performance with state-of-the-arts, which means it can also process multi-oriented text well.

\begin{table}[!htbp]
\begin{center}
\begin{tabular}{|c|c|c|c|}
\hline
{\bf Method} & {\bf Recall} & {\bf Precision} & {\bf Hmean}\\
\hline
SegLink~\cite{Authors_seglink} & 76.8 & 73.1 & 75.0 \\
\hline
RRPN~\cite{Authors_RRPN} & 77.0 & 84.0 & 80.0 \\
\hline
He \etal~\cite{Authors_deepdirect}& 81.0 & 92.0 & 86.0 \\
\hline
R2CNN~\cite{Authors_R2CNN} & 79.7 & 85.6 & 82.5 \\
\hline
TextSnake~\cite{Authors_TextSnake} & 80.4 & 84.9 & 82.6 \\
\hline
PixelLink~\cite{Authors_pixelink} & 82.0 & 85.5 & 83.7 \\
\hline
InceptText~\cite{Authors_Incepttext} & 80.6 & 90.5 & 85.3 \\
\hline
Mask Textspotter~\cite{Authors_MaskTextspotter} & 81.0 & {\bf 91.6} & 86.0 \\
\hline
{\bf Proposed} & {\bf 86.0} & 89.2 & 87.6 \\
\hline
FOTS~\cite{Authors_FOTS} & - & - & {\bf 88.0}\\
\hline
\end{tabular}
\end{center}
\caption{Results on ICDAR2015.}
\label{table6}
\end{table}

Table~\ref{table7} shows the results on MSRA-TD500 dataset and it shows that our detection method can support long straight text line detection and Chinese+English detection well. It achieves Hmean of 83.6\% and is better than all other methods.

\begin{table}[!htbp]
\begin{center}
\begin{tabular}{|c|c|c|c|}
\hline
{\bf Method} & {\bf Recall} & {\bf Precision} & {\bf Hmean}\\
\hline
EAST~\cite{Authors_EAST} & 67.4 & 87.3 & 76.1 \\
\hline
SegLink~\cite{Authors_seglink} & 70.0 & 86.0 & 77.0 \\
\hline
PixelLink~\cite{Authors_pixelink} & 73.2 & 83.0 & 77.8 \\
\hline
TextSnake~\cite{Authors_TextSnake} & 73.9 & 83.2 & 78.3 \\
\hline
InceptText~\cite{Authors_Incepttext} & 79.0 &  87.5 & 83.0 \\
\hline
MCN~\cite{Authors_MCNet} & 79.0 & {\bf 88.0} & 83.0 \\
\hline
{\bf Proposed} & {\bf 82.1} & 85.2 & {\bf 83.6} \\
\hline
\end{tabular}
\end{center}
\caption{Results on MSRA-TD500.}
\label{table7}
\end{table}

\subsection{Speed}
The speed of the proposed method is compared with two other methods as shown in Table~\ref{table_9}, which are all able to deal with arbitrary shape scene text. From the results, we can see that the speed of the proposed method is much faster than the other two methods. While pixel-wise prediction is needed in Mask Textspotter and TextSnake, it is not needed in the proposed method and less computation is needed.

\begin{table}[!htbp]
\begin{center}
\begin{tabular}{|c|c|c|}
\hline
{\bf Method} & {\bf Scale} & {\bf Speed}\\
\hline
TextSnake~\cite{Authors_TextSnake} & 768 & 1.1 fps \\
\hline
Mask Textspotter~\cite{Authors_MaskTextspotter} & 720 & 6.9 fps \\
\hline
{\bf Proposed} & 720 & {\bf 10.0 fps} \\
\hline
\end{tabular}
\end{center}
\caption{Speed compared on different detection methods supporting arbitrary shape texts.}
\label{table_9}
\end{table}

\subsection{Qualitative results}
\begin{figure}[!htbp]
\begin{center}
\includegraphics[width=\linewidth]{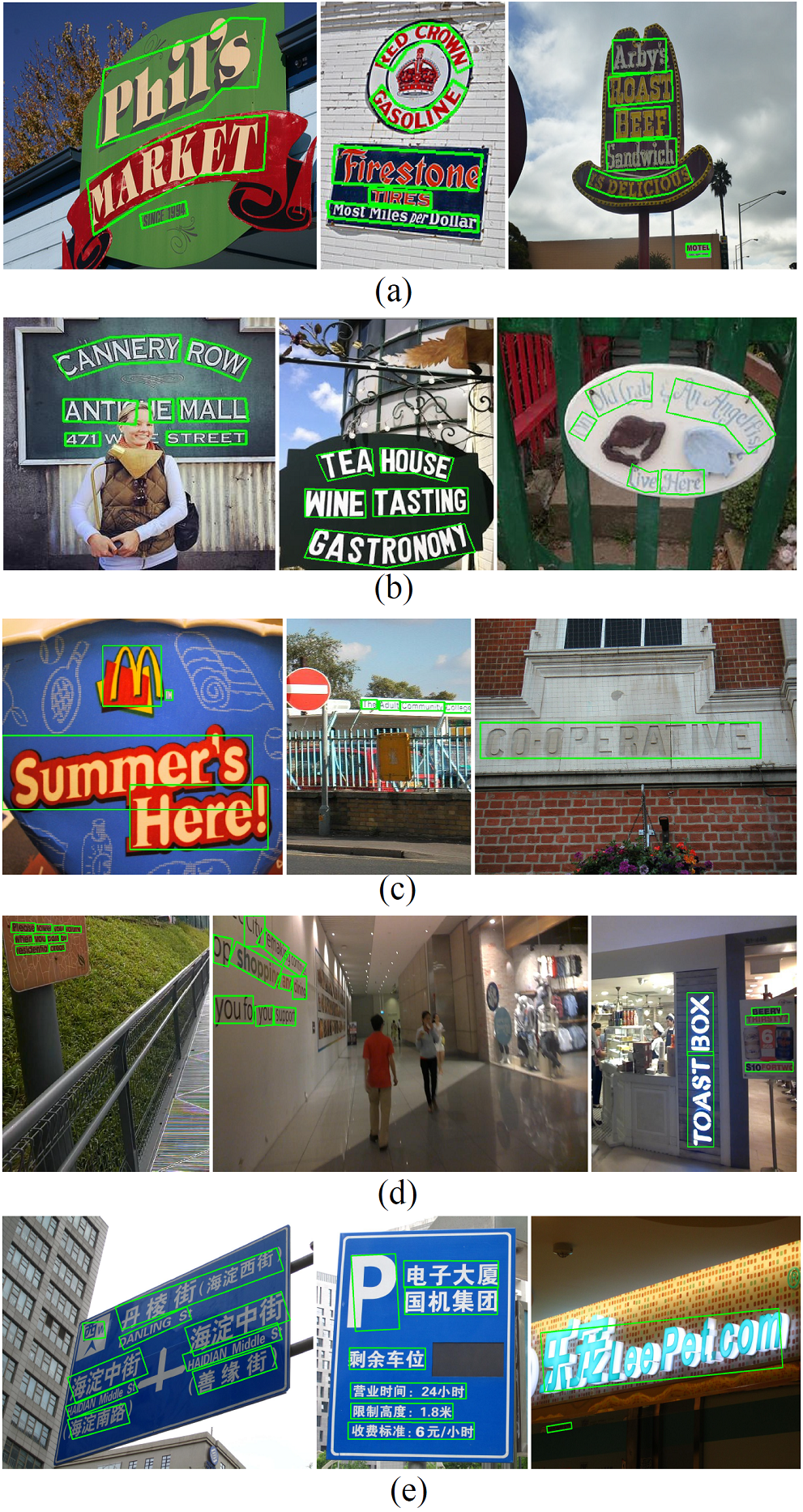}
\end{center}
   \caption{Results on different datasets. (a) results on CTW1500; (b) results on TotalText; (c) results on ICDAR2013; (d) results on ICDAR2015; (e) results on MSRA-TD500.}
\label{fig5}
\end{figure}

Figure~\ref{fig5} illustrates qualitative results on CTW1500, To-talText, ICDAR2013, ICDAR2015 and MSRA-TD500. It shows that the proposed method can deal with various texts of arbitrarily oriented or curved, different languages, non-uniform illuminations and different text lengths at word level or sentence level.

\section{Conclusion}

In this paper, we propose a robust arbitrary shape scene text detection method with adaptive text region representation. After text proposal using a Text-RPN, each text region is verified and refined using a RNN for predicting adaptive number of boundary points. Experiments on five benchmarks show that the proposed method can not only detect horizontal and oriented scene texts but also work well for arbitrary shape scene texts. Particularly, it outperforms existing methods significantly on CTW1500 and MSRA-TD500, which are typical of curved texts and multi-oriented texts, respectively.

In the future, the proposed method can be improved in several aspects. First, arbitrary shape scene text detection may can be improved by using corner point detection. This will require easier annotations for training images. Second, to fulfill the final goal of text recognition end-to-end text recognition for arbitrary shape scene text will be considered.

{\small
\bibliographystyle{ieee_fullname}
\bibliography{egbib}
}

\end{document}